\documentclass[11pt,a4paper]{article}
\usepackage[hyperref]{acl2019}
\usepackage{times}
\usepackage{latexsym}

\usepackage{url}

\usepackage{amsmath}
\usepackage{amssymb}
\usepackage{booktabs}
\usepackage{subcaption}
\usepackage{multirow}
\usepackage{tikz}
\usepackage{xcolor}

\usepackage[linesnumbered,ruled]{algorithm2e}

\aclfinalcopy 
\def\aclpaperid{2001} 

\newcommand\askip{\hspace{2pt}}
\newcommand\mailskip{\hspace{8pt}}

\title{Learning to Select, Track, and Generate for Data-to-Text}

\author{
    Hayate Iso{${}^{\thanks{~~Work was done during the internship at Artificial Intelligence Research Center, AIST}\dag}$}\askip
    Yui Uehara{${}^{\ddag}$}\askip
    Tatsuya Ishigaki{${}^{\natural\ddag}$}\askip
    Hiroshi Noji{${}^{\ddag}$}\askip
	\smallskip 
	~\\
	\bf{
	Eiji Aramaki{${}^{\dag\ddag}$}
	Ichiro Kobayashi{${}^{\flat\ddag}$}\askip
	Yusuke Miyao{${}^{\sharp\ddag}$}\askip
	Naoaki Okazaki{${}^{\natural\ddag}$}\askip
	Hiroya Takamura{${}^{\natural\ddag}$}} \\
	\smallskip 
	{${}^{\dag}$}Nara Institute of Science and Technology\askip
    {${}^{\ddag}$}Artificial Intelligence Research Center, AIST\askip \\
    {${}^\natural$}Tokyo Institute of Technology\askip 
    {${}^\flat$}Ochanomizu University\askip 
    {${}^\sharp$}The University of Tokyo\\
    {\tt
    \{iso.hayate.id3,aramaki\}@is.naist.jp\mailskip
    koba@is.ocha.ac.jp}\\
    {\tt
    \{yui.uehara,ishigaki.t,hiroshi.noji,takamura.hiroya\}@aist.go.jp
    }\\
    {\tt
    yusuke@is.s.u-tokyo.ac.jp\mailskip
    okazaki@c.titech.ac.jp 
  }
}

\date{}

\begin{document}
\maketitle
\begin{abstract}
We propose a data-to-text generation model with two modules, one for tracking and the other for text generation.
Our tracking module selects and keeps track of salient information and memorizes which record has been mentioned.
Our generation module generates a summary conditioned on the state of tracking module.
Our model is considered to simulate the human-like writing process that gradually selects the information by determining the intermediate variables while writing the summary.
In addition, we also explore the effectiveness of the writer information for generation.
Experimental results show that our model outperforms existing models in all evaluation metrics even without writer information.
Incorporating writer information further improves the performance, contributing to content planning and surface realization. 
\end{abstract}

\section{Introduction}
Advances in sensor and data storage technologies have rapidly increased the amount of data produced in various fields such as weather, finance, and sports.
In order to address the information overload caused by the massive data, data-to-text generation technology, which expresses the contents of data in natural language, becomes more important~\cite{barzilay2005collective}.
Recently, neural methods can generate high-quality short summaries especially from small pieces of data~\cite{liu2018table}. 

Despite this success, it remains challenging to generate a high-quality long summary from data~\cite{wiseman2017challenges}.
One reason for the difficulty is because the input data is too large for a naive model to find its salient part, i.e., to determine which part of the data should be mentioned. 
In addition, the salient part moves as the summary explains the data.
For example, when generating a summary of a basketball game (Table~\ref{tab:example} (b)) from the box score (Table~\ref{tab:example} (a)), 
the input contains numerous data records about the game: e.g., \textit{Jordan Clarkson scored 18 points}.
Existing models often refer to the same data record multiple times~\cite{puduppully2019data}. The models may mention an incorrect data record, e.g., \textit{Kawhi Leonard added 19 points}: the summary should mention \textit{LaMarcus Aldridge}, who scored 19 points.
Thus, we need a model that finds salient parts, tracks transitions of salient parts,
and expresses information faithful to the input.

In this paper, we propose a novel data-to-text generation model with two modules, one for saliency tracking and another for text generation.
The tracking module keeps track of saliency in the input data: when the module detects a saliency transition, the tracking module selects a new data record\footnote{We use `data record' and `relation' interchangeably.} and updates the state of the tracking module.
The text generation module generates a document conditioned on the current tracking state.
Our model is considered to imitate the human-like writing process that gradually selects and tracks the data while generating the summary.
In addition, we note some writer-specific patterns and characteristics: how data records are selected to be mentioned; and how data records are expressed as text, e.g., the order of data records and the word usages. We also incorporate writer information into our model.

The experimental results demonstrate that, even without writer information, our model achieves the best performance among the previous models in all evaluation metrics: 94.38\% precision of relation generation, 42.40\% F1 score of content selection, 19.38\% normalized Damerau-Levenshtein Distance (DLD) of content ordering, and 16.15\% of \textsc{BLEU} score. We also confirm that adding writer information further improves the performance.

\begin{table*}[t]{\scriptsize
    \centering
    \begin{subtable}{0.52\linewidth}
        \centering
        \begin{tabular}{@{}l@{~~}c@{~~}c@{~~}c@{~~}c@{~~}c@{~~}c@{~~}c@{~~}c@{~~}l@{}}
        \toprule
        \textsc{Team} & \textsc{H/V} & \textsc{Win} &\textsc{Loss} &\textsc{Pts} & \textsc{Reb} & \textsc{Ast} & \textsc{Fg\_Pct} & \textsc{Fg3\_Pct} &\textsc{$\dots$} \\
        \midrule
        \textsc{Knicks} & H &{\color[HTML]{2ca02c}\textbf{16}} &{\color[HTML]{2ca02c}\textbf{19}} & {\color[HTML]{2ca02c}\textbf{104}} &\textsc{46} &\textsc{26} & \textsc{45} & \textsc{46} &\textsc{$\dots$} \\ 
        \textsc{Bucks} & V &{\color[HTML]{2ca02c}\textbf{18}} &{\color[HTML]{2ca02c}\textbf{16}} &{\color[HTML]{2ca02c}\textbf{105}} & \textsc{42} & \textsc{20} & \textsc{47} & \textsc{32} & \textsc{$\dots$} \\ 
        \bottomrule
        \end{tabular}
        \vspace{1em}\\
        \begin{tabular}{@{}l@{~~}c@{~~}c@{~}c@{~~}c@{~~}c@{~~}c@{~~}c@{~~}c@{~~}l@{}}
        \toprule
        \textsc{Player} & \textsc{H/V} & \textsc{Pts} & \textsc{Reb} & \textsc{Ast} & \textsc{Blk} & \textsc{Stl} & \textsc{Min} & \textsc{City} & \textsc{$\dots$}\\
        \midrule
        \textsc{Carmelo Anthony} & \textsc{H}& {\color[HTML]{2ca02c}\textbf{30}} & {\color[HTML]{2ca02c}\textbf{11}} & {\color[HTML]{2ca02c}\textbf{7}} & \textsc{0} & {\color[HTML]{2ca02c}\textbf{2}} & \textsc{37} & \textsc{New York} & \textsc{$\dots$}\\
        \textsc{Derrick Rose} & \textsc{H}& {\color[HTML]{2ca02c}\textbf{15}} & {\color[HTML]{2ca02c}\textbf{3}} & {\color[HTML]{2ca02c}\textbf{4}} & \textsc{0} & \textsc{1} & \textsc{33} & \textsc{New York} & \textsc{$\dots$}\\
        \textsc{Courtney Lee} & \textsc{H}& {\color[HTML]{2ca02c}\textbf{11}} & {\color[HTML]{2ca02c}\textbf{2}} & {\color[HTML]{2ca02c}\textbf{3}} & \textsc{1} & \textsc{1} & \textsc{38} & \textsc{New York} & \textsc{$\dots$}\\
        \textsc{Giannis Antetokounmpo} & \textsc{V} & {\color[HTML]{2ca02c}\textbf{27}} & {\color[HTML]{2ca02c}\textbf{13}} & {\color[HTML]{2ca02c}\textbf{4}} & {\color[HTML]{2ca02c}\textbf{3}} & \textsc{1} & \textsc{39} & \textsc{Milwaukee} & \textsc{$\dots$}\\
        \textsc{Greg Monroe} & \textsc{V} & {\color[HTML]{2ca02c}\textbf{18}} & {\color[HTML]{2ca02c}\textbf{9}} & {\color[HTML]{2ca02c}\textbf{4}} & \textsc{1} & {\color[HTML]{2ca02c}\textbf{3}} & \textsc{31} & \textsc{Milwaukee} & \textsc{$\dots$}\\
        \textsc{Jabari Parker} & \textsc{V}& {\color[HTML]{2ca02c}\textbf{15}} & {\color[HTML]{2ca02c}\textbf{4}} & {\color[HTML]{2ca02c}\textbf{3}} & \textsc{0} & \textsc{1} & \textsc{37} &\textsc{Milwaukee} & \textsc{$\dots$}\\
        \textsc{Malcolm Brogdon} & \textsc{V}& {\color[HTML]{2ca02c}\textbf{12}} & {\color[HTML]{2ca02c}\textbf{6}} & {\color[HTML]{2ca02c}\textbf{8}} & \textsc{0} & \textsc{0} & \textsc{38} &\textsc{Milwaukee} & \textsc{$\dots$}\\
        \textsc{Mirza Teletovic} & \textsc{V}& {\color[HTML]{2ca02c}\textbf{13}} & \textsc{1} & \textsc{0} & \textsc{0} & \textsc{0} & \textsc{21} &\textsc{Milwaukee} & \textsc{$\dots$}\\
        \textsc{John Henson} & \textsc{V}& \textsc{2} & \textsc{2} & \textsc{0} & \textsc{0} & \textsc{0} & \textsc{14} & \textsc{Milwaukee} & \textsc{$\dots$}\\
        \textsc{$\dots$} & \textsc{$\dots$} & \textsc{$\dots$} & \textsc{$\dots$} & \textsc{$\dots$}& \textsc{$\dots$} & \textsc{$\dots$}\\
        \bottomrule
    \end{tabular}
    \caption{Box score: Top contingency table shows number of wins and losses and summary of each game. Bottom table shows statistics of each player such as points scored (\textsc{Player}'s \textsc{Pts}), and total rebounds (\textsc{Player}'s \textsc{Reb}).}
    \label{sub:box}
    \end{subtable}~\hspace{2em}
    \begin{subtable}{0.43\linewidth}
        \centering
        \begin{tikzpicture}
        \node [rectangle, draw, thick, fill=blue!0, text width=0.9\linewidth,  rounded corners, inner sep = 8pt, minimum height=5em]{
        The \textbf{Milwaukee Bucks} defeated the \textbf{New York Knicks}, {\color[HTML]{2ca02c}\textbf{105}}-{\color[HTML]{2ca02c}\textbf{104}}, at Madison Square Garden on Wednesday. The \textbf{Knicks} ({\color[HTML]{2ca02c}\textbf{16}}-{\color[HTML]{2ca02c}\textbf{19}}) checked in to Wednesday's contest looking to snap a five-game losing streak and heading into the fourth quarter, they looked like they were well on their way to that goal. $\dots$ \textbf{Antetokounmpo} led the Bucks with {\color[HTML]{2ca02c}\textbf{27}} points, {\color[HTML]{2ca02c}\textbf{13}} rebounds, {\color[HTML]{2ca02c}\textbf{four}} assists, a steal and {\color[HTML]{2ca02c}\textbf{three}} blocks, his second consecutive double-double. \textbf{Greg Monroe} actually checked in as the second-leading scorer and did so in his customary bench role, posting {\color[HTML]{2ca02c}\textbf{18}} points, along with {\color[HTML]{2ca02c}\textbf{nine}} boards, {\color[HTML]{2ca02c}\textbf{four}} assists, {\color[HTML]{2ca02c}\textbf{three}} steals and a block. \textbf{Jabari Parker} contributed {\color[HTML]{2ca02c}\textbf{15}} points, {\color[HTML]{2ca02c}\textbf{four}} rebounds, {\color[HTML]{2ca02c}\textbf{three}} assists and a steal.  \textbf{Malcolm Brogdon} went for {\color[HTML]{2ca02c}\textbf{12}} points, {\color[HTML]{2ca02c}\textbf{eight}} assists and {\color[HTML]{2ca02c}\textbf{six}} rebounds. \textbf{Mirza Teletovic} was productive in a reserve role as well, generating {\color[HTML]{2ca02c}\textbf{13}} points and a rebound. $\dots$ \textbf{Courtney Lee} checked in with {\color[HTML]{2ca02c}\textbf{11}} points, {\color[HTML]{2ca02c}\textbf{three}} assists, {\color[HTML]{2ca02c}\textbf{two}} rebounds, a steal and a block. $\dots$ The Bucks and Knicks face off once again in the second game of the home-and-home series, with the meeting taking place Friday night in Milwaukee.
        };
        \end{tikzpicture}
        \caption{NBA basketball game summary: Each summary consists of game victory or defeat of the game and highlights of valuable players.}
        \label{sub:summary}
    \end{subtable}
    \caption{Example of input and output data: task defines box score (\ref{sub:box}) used for input and summary document of  game (\ref{sub:summary}) used as output. 
    Extracted entities are shown in \textbf{bold face}. Extracted values are shown in {\color[HTML]{2ca02c} green}.
    }
    \label{tab:example}
    }
\end{table*}

\section{Related Work}
\subsection{Data-to-Text Generation}
Data-to-text generation is a task for generating descriptions from structured or non-structured data including sports commentary~\cite{tanaka1998reactive,chen2008learning,taniguchi2019generating}, weather forecast~\cite{liang2009learning,mei2016talk}, biographical text from infobox in Wikipedia~\cite{lebret2016neural,sha2018order,liu2018table} and market comments from stock prices~\cite{murakami2017learning,aoki2018generating}.

Neural generation methods have become the mainstream approach for data-to-text generation.
The encoder-decoder framework~\cite{cho2014learning,sutskever2014sequence} with the attention~\cite{bahdanau2015neural,luong2015effective} and copy mechanism~\cite{gu2016incorporating,gulcehre2016pointing} has successfully applied to data-to-text tasks.
However, neural generation methods sometimes yield fluent but inadequate descriptions~\cite{tu2017context}.
In data-to-text generation, descriptions inconsistent to the input data are problematic.

Recently, \citet{wiseman2017challenges} introduced the \textsc{RotoWire} dataset, which contains multi-sentence summaries of basketball games with box-score (Table~\ref{tab:example}). This dataset requires the selection of a salient subset of data records for generating descriptions.
They also proposed automatic evaluation metrics for
measuring the informativeness of generated summaries.

\citet{puduppully2019data} proposed a two-stage method 
that first predicts the sequence of data records to be mentioned
and then generates a summary conditioned on the
predicted sequences. 
Their idea is similar to ours in that the both consider a sequence of data records as content planning.
However, our proposal differs from theirs in that ours uses a recurrent neural network for saliency tracking, and that our decoder dynamically chooses a data record to be mentioned without fixing a sequence of data records.

\subsection{Memory modules}
The memory network can be used to maintain and update representations of the salient information~\cite{weston2015memory,sukhbaatar2015end,graves2016hybrid}.
This module is often used in natural language understanding to keep track of the entity state~\cite{kobayashi2016dynamic,hoang2018entity,bosselut2018simulating}.

Recently, entity tracking has been popular for generating coherent text~\cite{kiddon2016globally,ji2017dynamic,yang2017reference,clark2018neural}.
\citet{kiddon2016globally} proposed a neural checklist model that updates predefined item states.
\citet{ji2017dynamic} proposed an entity representation for the language model. Updating entity tracking states when the entity is introduced, their method selects the salient entity state.

Our model extends this entity tracking module for data-to-text generation tasks.
The entity tracking module selects the salient entity and appropriate attribute in each timestep, updates their states, and generates coherent summaries from the selected data record.

\begin{table*}[t]
    \scriptsize
    \centering
    \addtolength{\tabcolsep}{-1pt}
    \begin{tabular}{c|ccccccccccccccccc}
        \toprule
        $t$ & 199 & 200 & 201 & 202 & 203 & 204 & 205 & 206 & 207 & 208 & 209\\
        \midrule
         $Y_{t}$& Jabari & Parker & contributed & 15 & points & , & four & rebounds & , & three & assists\\
         \midrule
         $Z_{t}$& 1 & 1 & 0 & 1 & 0 & 0 & 1 & 0 & 0 & 1 & 0 \\
         \multirow{2}{*}{$E_{t}$} & \textsc{Jabari} & \textsc{Jabari} & \multirow{2}{*}{-} & \textsc{Jabari} & \multirow{2}{*}{-} & \multirow{2}{*}{-} & \textsc{Jabari} & \multirow{2}{*}{-} & \multirow{2}{*}{-} & \textsc{Jabari} & \multirow{2}{*}{-}\\
         &\textsc{Parker} & \textsc{Parker} & & \textsc{Parker} & & & \textsc{Parker} & & & \textsc{Parker} & & \\
         $A_{t}$& \textsc{First Name} & \textsc{Last Name} & - & \textsc{Player Pts} & - & - & \textsc{Player Reb} & - & - & \textsc{Player Ast} & - \\
         $N_{t}$& - & - & - & 0 & - & - & 1 & - & - & 1 & - \\
         \bottomrule
    \end{tabular}
    
    \caption{Running example of our model's generation process. At every time step $t$, model predicts each random variable. Model firstly determines whether to refer to data records ($Z_{t} = 1$) or not ($Z_{t} = 0$). If random variable $Z_t = 1$, model selects entity $E_t$, its attribute $A_t$ and binary variables $N_t$ if needed.
    For example, at $t = 202$, model predicts random variable $Z_{202} = 1$ and then selects the entity \textsc{\textbf{Jabari Parker}} and its attribute \textsc{\textbf{Player Pts}}. Given these values, model outputs token $\mathbf{15}$ from selected data record.
    }
    \label{tab:annotate}
    \vspace{-1em}
\end{table*}

\section{Data}
\label{section:data}
Through careful examination, we found that in the original dataset \textsc{RotoWire}, some NBA games have two documents, one of which is sometimes in the training data and the other is in the test or validation data. Such documents are similar to each other, though not identical. To make this dataset more reliable as an experimental dataset, we created a new version.

We ran the script provided by \citet{wiseman2017challenges}, which is for crawling the \textsc{RotoWire} website for NBA game summaries. The script collected approximately 78\% of the documents in the original dataset; the remaining documents disappeared. We also collected the box-scores associated with the collected documents. We observed that some of the box-scores were modified compared with the original \textsc{RotoWire} dataset.

The collected dataset contains 3,752 instances (i.e., pairs of a document and box-scores). However, the four shortest documents were not summaries; they were, for example, an announcement about the postponement of a match. We thus deleted these 4 instances and were left with 3,748 instances.
We followed the dataset split by \citet{wiseman2017challenges} to split our dataset into training, development, and test data. We found 14 instances that didn't have corresponding instances in the original data. We randomly classified 9, 2, and 3 of those 14 instances respectively into training, development, and test data. Finally, the sizes of our training, development, test dataset are respectively 2,714, 534, and 500. On average, each summary has 384 tokens and 644 data records. Each match has only one summary in our dataset, as far as we checked. 
We also collected the writer of each document. 
Our dataset contains 32 different writers.
The most prolific writer in our dataset wrote 607 documents. There are also writers who wrote less than ten documents. On average, each writer wrote 117 documents.
We call our new dataset \textsc{RotoWire-Modified}.\footnote{For information about the dataset, please follow this link: \url{https://github.com/aistairc/rotowire-modified}}

\section{Saliency-Aware Text Generation}
At the core of our model is a neural language model with a memory state $\boldsymbol{h}^{\textsc{LM}}$ to
generate a summary $y_{1:T} = (y_1, \dots, y_T)$ given a set of data records $\boldsymbol{x}$.
Our model has another memory state $\boldsymbol{h}^{\textsc{Ent}}$, which is used to remember the data records that have been referred to.
$\boldsymbol{h}^{\textsc{Ent}}$ is also used to update $\boldsymbol{h}^{\textsc{LM}}$, meaning that the referred data records affect the text generation. 

Our model decides whether to refer to $\boldsymbol{x}$, which data record $r\in \boldsymbol{x}$ to be mentioned, and how to express a number. The selected data record is used to update $\boldsymbol{h}^{\textsc{Ent}}$.
Formally, we use the four variables:
\vspace{-1.0em}
\begin{enumerate}
    \setlength{\itemsep}{0pt}
    \setlength{\parskip}{0pt}
    \setlength{\leftskip}{-4mm}
    \item $ Z_t $: binary variable that determines whether the model refers to input $\boldsymbol{x}$ at time step $t$ ($ Z_t = 1 $).
    \item $ E_t $: At each time step $t$, this variable indicates the salient entity (e.g., \textsc{Hawks}, \textsc{LeBron James}).
    \item $ A_t $: At each time step $t$, this variable indicates the salient attribute to be mentioned (e.g., \textsc{Pts}).
    \item $ N_t $: If attribute $A_t$ of the salient entity $E_t$ is a numeric attribute, this variable determines if a value in the data records should be 
    output in Arabic numerals (e.g., 50) or in English words (e.g., five).
\end{enumerate}
\vspace{-0.5em}

To keep track of the salient entity, our model predicts these random variables at each time step $t$ through its summary generation process. Running example of our model is shown in Table~\ref{tab:annotate} and full algorithm is described in Appendix~\ref{sec:algorithm}.
In the following subsections, we explain how to initialize the model, predict these random variables, and generate a summary.
Due to space limitations, bias vectors are omitted.

Before explaining our method, we describe our notation. Let $\mathcal{E}$ and $\mathcal{A}$ denote the sets of entities and attributes, respectively.
Each record $r \in \boldsymbol{x}$ consists of entity $e\in\mathcal{E}$, attribute $a\in\mathcal{A}$, and its value $\boldsymbol{x}[e, a]$, and is therefore represented as $r = (e, a, \boldsymbol{x}[e, a])$.
For example, the box-score in Table~\ref{tab:example} has a record $r$ such that $e = \textsc{Anthony Davis}, a = \textsc{Pts},$ and $\boldsymbol{x}[e, a] = 20$.

\subsection{Initialization}
\label{sec:init}
Let $\boldsymbol{r}$ denote the embedding of data record $r \in \boldsymbol{x}$. 
Let $\bar{\boldsymbol{e}}$ denote the embedding of entity $e$. Note that $\bar{\boldsymbol{e}}$ depends on the set of data records, i.e., it depends on the game.
We also use $\boldsymbol{e}$ for static embedding of entity $e$, which, on the other hand, does not depend on the game.

Given the embedding of entity $\boldsymbol{e}$, attribute $\boldsymbol{a}$, and its value $\boldsymbol{v}$, we use the concatenation layer to combine the information from these vectors to produce the embedding of each data record $(e,a,v)$, denoted as $\boldsymbol{r}_{e,a,v}$ as follows:
\begin{align}
    \boldsymbol{r}_{e,a,v} = \tanh\left(\boldsymbol{W}^{\textsc{R}}(\boldsymbol{e} \oplus \boldsymbol{a} \oplus \boldsymbol{v})\right),
    \label{init}
\end{align}
where $\oplus$ indicates the concatenation of vectors, and $\boldsymbol{W}^{\textsc{R}}$ denotes a weight matrix.\footnote{We also concatenate the embedding vectors that represents whether the entity is in home or away team.}

We obtain $\bar{\boldsymbol{e}}$ in the set of data records $\boldsymbol{x}$,
by summing all the data-record embeddings transformed by a matrix:
\begin{align}
    \bar{\boldsymbol{e}} = \tanh\left(\sum_{a\in \mathcal{A}} \boldsymbol{W}^{\textsc{A}}_{a}\boldsymbol{r}_{e, a, \boldsymbol{x}[e,a]}\right),
\end{align}
where $\boldsymbol{W}^{\textsc{A}}_{a}$ is a weight matrix for attribute $a$. 
Since $\bar{\boldsymbol{e}}$ depends on the game as above, $\bar{\boldsymbol{e}}$ is supposed to represent how entity $e$ played in the game.

To initialize the hidden state of each module, we use embeddings of $<$\textsc{SoD}$>$ for $\boldsymbol{h}^{\textsc{LM}}$ and averaged embeddings of $\bar{\boldsymbol{e}}$ for $\boldsymbol{h}^{\textsc{ENT}}$.

\subsection {Saliency transition}\label{subsec:saliency}
Generally, the saliency of text changes during text generation.
In our work, we suppose that the saliency is represented as the entity and its attribute being talked about. We therefore propose a model that refers to a data record at each timepoint, and transitions to another as text goes.

To determine whether to transition to another data record or not at time $t$, the model calculates the following probability:
\begin{align}
    p(Z_t = 1 \mid \boldsymbol{h}_{t-1}^{\textsc{LM}}, \boldsymbol{h}_{t-1}^{\textsc{Ent}}) = \sigma(\boldsymbol{W}_{z}(\boldsymbol{h}_{t-1}^{\textsc{LM}} \oplus \boldsymbol{h}_{t-1}^{\textsc{Ent}})),
\end{align}
where $\sigma (\cdot) $ is the sigmoid function.
If $p(Z_t = 1 \mid \boldsymbol{h}_{t-1}^{\textsc{LM}}, \boldsymbol{h}_{t-1}^{\textsc{Ent}})$ is high, the model transitions to another data record.

When the model decides to transition to another, the model then determines which entity and attribute to refer to, and generates the next word (Section~\ref{sec:ent}).
On the other hand, if the model decides not transition to another, the model generates the next word without updating the tracking states $\boldsymbol{h}^{\textsc{Ent}}_t = \boldsymbol{h}^{\textsc{Ent}}_{t-1}$ (Section~\ref{sec:out}).

\subsection{Selection and tracking}
\label{sec:ent}
When the model refers to a new data record ($Z_t = 1$), it selects an entity and its attribute. It also tracks the saliency by putting the information about the selected entity and attribute into the memory vector $\boldsymbol{h}^{\textsc{Ent}}$.
The model begins to select the subject entity and update the memory states if the subject entity will change. 

Specifically, the model first calculates the probability of selecting an entity:
\begin{align}
    &p(E_t = e \mid  \boldsymbol{h}_{t-1}^{\textsc{LM}}, \boldsymbol{h}_{t-1}^{\textsc{Ent}})\nonumber\\
    \propto
    &\begin{cases}
        \exp\left({\boldsymbol{h}^{\textsc{Ent}}_s\boldsymbol{W}^{\textsc{Old}}\boldsymbol{h}_{t-1}^{\textsc{LM}}}\right) & \text{if } e \in \mathcal{E}_{t-1}\\
        \exp\left({\bar{\boldsymbol{e}}\boldsymbol{W}^{\textsc{New}}\boldsymbol{h}_{t-1}^{\textsc{LM}}}\right) & \text{otherwise}
    \end{cases},
\end{align}
where 
$\mathcal{E}_{t-1}$ is the set of entities that have already been referred to by time step $t$,
and $s$ is defined as $ s = {\max \{s: s \leq t - 1, e = e_s \}} $, which indicates the time step when this entity was last mentioned.

The model selects the most probable entity as the next salient entity and updates the set of entities that appeared ($\mathcal{E}_t = \mathcal{E}_{t - 1} \cup \{e_t\}$).

If the salient entity changes $(e_t \not= e_{t - 1})$, the model updates the hidden state of the tracking model $\boldsymbol{h}^{\textsc{Ent}}$ with a recurrent neural network with a gated recurrent unit~\citep[\textsc{Gru};][]{chung2014empirical}:
\begin{align}
     \boldsymbol{h}_{t}^{\textsc{Ent}'} 
    = 
    \begin{cases}
    \boldsymbol{h}_{t-1}^{\textsc{Ent}} \hspace{2.8cm}\text{ if } e_t = e_{t-1}\\
    \textsc{Gru}^{\textsc{E}}(\bar{\boldsymbol{e}}, \boldsymbol{h}_{t-1}^{\textsc{Ent}}) \hspace{0.4cm}  \text{ else if } e_t \not \in \mathcal{E}_{t-1}\\
    \textsc{Gru}^{\textsc{E}}(\boldsymbol{W}^\textsc{S}_s\boldsymbol{h}_s^{\textsc{Ent}}, \boldsymbol{h}_{t-1}^{\textsc{Ent}}) \hspace{0.2cm} \text{  otherwise.}
    \end{cases}
\end{align}
Note that if the selected entity at time step $t$, $e_t$, is identical to the previously selected entity $e_{t-1}$, the hidden state of the tracking model is not updated.

If the selected entity $e_t$ is new ($e_t \not \in \mathcal{E}_{t-1}$), the hidden state of the tracking model is updated with the embedding $\bar{\boldsymbol{e}}$ of entity $e_t$ as input.
In contrast, if entity $e_t$ has already appeared in the past ($e_t \in \mathcal{E}_{t-1}$) but is not identical to the previous one $(e_t \not= e_{t-1})$, we use  $\boldsymbol{h}_s^{\textsc{Ent}}$ (i.e., the memory state when this entity last appeared) 
to fully exploit the local history of this entity.

Given the updated hidden state of the tracking model $\boldsymbol{h}_{t}^{\textsc{Ent}}$, we next select the attribute of the salient entity by the following probability:
\begin{align}
    &p(A_t = a \mid e_t,\boldsymbol{h}_{t-1}^{\textsc{LM}}, \boldsymbol{h}_{t}^{\textsc{Ent}'}) \\
    \propto &\exp\left(\boldsymbol{r}_{e_t, a, \boldsymbol{x}[e_t,a]}\boldsymbol{W}^{\textsc{Attr}}  (\boldsymbol{h}_{t-1}^{\textsc{LM}} \oplus \boldsymbol{h}_{t}^{\textsc{Ent}'})\right).\nonumber
\end{align}
After selecting $a_t$, i.e., the most probable attribute of the salient entity, the tracking model updates the memory state $\boldsymbol{h}_{t}^{\textsc{Ent}}$ with the embedding of the data record $\boldsymbol{r}_{e_t, a_t, \boldsymbol{x}[e_t, a_t]}$ introduced in Section~\ref{sec:init}:
\begin{align}
    \boldsymbol{h}_{t}^{\textsc{Ent}} = \textsc{Gru}^{\textsc{A}}(\boldsymbol{r}_{e_t, a_t, \boldsymbol{x}[e_t,a_t]}, \boldsymbol{h}_{t}^{\textsc{Ent}'}).
\end{align}

\subsection{Summary generation}
\label{sec:out}
Given two hidden states, one for language model $\boldsymbol{h}_{t-1}^{\textsc{LM}}$ and the other for tracking model $\boldsymbol{h}_{t}^{\textsc{Ent}}$, the model generates the next word $y_t$.
We also incorporate a copy mechanism that copies the value of the salient data record $\boldsymbol{x}[e_t, a_t]$.

If the model refers to a new data record ($Z_t = 1$), it directly copies the value of the data record $\boldsymbol{x}[e_t, a_t]$.
However, the values of numerical attributes 
can be expressed in at least two different manners: Arabic numerals (e.g., {\it 14}) and English words (e.g., {\it fourteen}).
We decide which one to use by the following probability:
\begin{align}
    p(N_t = 1 \mid \boldsymbol{h}_{t-1}^{\textsc{LM}}, \boldsymbol{h}_{t}^{\textsc{Ent}}) = \sigma(\boldsymbol{W}^{\textsc{N}}(\boldsymbol{h}_{t-1}^{\textsc{LM}}\oplus \boldsymbol{h}_{t}^{\textsc{Ent}})),
\end{align}
where $\boldsymbol{W}^{\textsc{N}}$ is a weight matrix.
The model then updates the hidden states of the language model:
\begin{align}
     \boldsymbol{h}_t'&= \tanh\left(\boldsymbol{W}^{\textsc{H}}(\boldsymbol{h}_{t-1}^{\textsc{LM}} \oplus\boldsymbol{h}_{t}^{\textsc{Ent}})\right),\label{eq:hidden}
\end{align}
where $\boldsymbol{W}^{\textsc{H}}$ is a weight matrix.

If the salient data record is the same as the previous one ($Z_t = 0$), it predicts the next word $y_t$ via a probability over words conditioned on the context vector $\boldsymbol{h}_t'$:
\begin{align}
    p(Y_t \mid \boldsymbol{h}_t') = \text{softmax}(\boldsymbol{W}^{\textsc{Y}} \boldsymbol{h}_t').\label{eq:prob}
\end{align}
Subsequently, the hidden state of language model $\boldsymbol{h}^{\textsc{LM}}$ is updated:
\begin{align}
    \boldsymbol{h}_t^{\textsc{LM}} &= \textsc{LSTM}(\boldsymbol{y}_{t}\oplus\boldsymbol{h}_t', \boldsymbol{h}_{t-1}^{\textsc{LM}}),
\end{align}
where $\boldsymbol{y}_t$ is the embedding of the word generated at time step $t$.\footnote{In our initial experiment, we observed a word repetition problem when the tracking model is not updated during generating each sentence. To avoid this problem, we also update the tracking model with special trainable vectors $\boldsymbol{v}_{\textsc{REFRESH}}$ to refresh these states after our model generates a period: $\boldsymbol{h}_t^{\textsc{Ent}} = \textsc{Gru}^{A}(\boldsymbol{v}_{\textsc{Refresh}}, \boldsymbol{h}_t^{\textsc{Ent}})$}

\subsection{Incorporating writer information}
\label{subsec:writer}
We also incorporate the information about the writer of the summaries into our model.
Specifically, 
instead of using Equation (\ref{eq:hidden}), 
we concatenate the embedding $\boldsymbol{w}$ of a writer
to $\boldsymbol{h}_{t-1}^{\textsc{LM}}\oplus\boldsymbol{h}_{t}^{\textsc{Ent}}$
to construct context vector $\boldsymbol{h}_t'$:
\begin{align}
    \boldsymbol{h}_t'&= \tanh\left(\boldsymbol{W}'^{\textsc{H}}(\boldsymbol{h}_{t-1}^{\textsc{LM}} \oplus\boldsymbol{h}_{t}^{\textsc{Ent}}\oplus \boldsymbol{w})\right),
 \end{align}
where $\boldsymbol{W}'^{\textsc{H}}$ is a new weight matrix.
Since this new context vector $\boldsymbol{h}_t'$ is used for calculating the probability over words in Equation~(\ref{eq:prob}), the writer information will directly affect word generation, which is regarded as surface realization in terms of traditional text generation. Simultaneously, context vector $\boldsymbol{h}_t'$ enhanced with the writer information is used to obtain $\boldsymbol{h}_t^{\textsc{LM}}$, which is the hidden state of the language model and is further used to select the salient entity and attribute, as mentioned in Sections~\ref{subsec:saliency} and \ref{sec:ent}. Therefore, in our model, the writer information affects both surface realization and content planning.
 
\subsection{Learning objective}
We apply fully supervised training that maximizes the following log-likelihood:
\begin{align*}
    &\log p(Y_{1:T}, Z_{1:T}, E_{1:T}, A_{1:T}, N_{1:T} \mid \boldsymbol{x})\\
    =&\sum_{t=1}^T\log p(Z_t = z_t \mid \boldsymbol{h}_{t-1}^{\textsc{LM}}, \boldsymbol{h}_{t-1}^{\textsc{Ent}})~\\
    + &\sum_{t:Z_t = 1}\log p(E_t = e_t \mid \boldsymbol{h}_{t-1}^{\textsc{LM}}, \boldsymbol{h}_{t-1}^{\textsc{Ent}})\\
    + &\sum_{t:Z_t = 1}\log p(A_t = a_t \mid e_t,  \boldsymbol{h}_{t-1}^{\textsc{LM}}, \boldsymbol{h}_{t}^{\textsc{Ent}'})\\
    +&\sum_{t:Z_t = 1, a_t \text{is num\_attr}} \log p(N_t = n_t \mid  \boldsymbol{h}_{t-1}^{\textsc{LM}}, \boldsymbol{h}_{t}^{\textsc{Ent}})\\
    + & \sum_{t: Z_t = 0}\log p(Y_{t} = y_t \mid \boldsymbol{h}_t')
\end{align*}

\begin{table*}[t]{
    \centering
    \addtolength{\tabcolsep}{-2pt}
    \begin{tabular}{r|cc|ccc|c|c}
        \toprule
        \multirow{2}{*}{Method} & \multicolumn{2}{c|}{\textsc{RG}} &\multicolumn{3}{c|}{\textsc{CS}} & \textsc{CO} & \multirow{2}{*}{\textsc{Bleu}}\\
        & \# & P\% & P\% & R\% & F1\% & DLD\%\\
        \midrule\midrule
        \textsc{Gold} & 27.36 & 93.42 & 100. & 100. & 100. & 100. & 100.\\
        \textsc{Templates} & 54.63 & 100. & 31.01 & 58.85 & 40.61 & 17.50 & 8.43 \\
        \midrule
        \citet{wiseman2017challenges} & 22.93 & 60.14 & 24.24 & 31.20 & 27.29 & 14.70 & 14.73\\
        \citet{puduppully2019data} & 33.06 & 83.17 & 33.06 & 43.59 & 37.60 & 16.97 & 13.96 \\
        \textsc{Proposed} & 39.05 & \textbf{94.43} & \textbf{35.77} & \textbf{52.05} & \textbf{42.40} & \textbf{19.38} & \textbf{16.15}\\
         \bottomrule
    \end{tabular}
    \caption{Experimental result. Each metric evaluates whether important information (CS) is described accurately (RG) and in correct order (CO).}
    \label{tab:result}
    }
\end{table*}

\section{Experiments}
\subsection{Experimental settings}

We used \textsc{RotoWire-Modified} as the dataset for our experiments, which we explained in Section~\ref{section:data}.
The training, development, and test data respectively contained 
2,714, 534, and 500 games.

Since we take a supervised training approach, we need the annotations of the random variables (i.e., $Z_t$, $E_t$, $A_t$, and $N_t$) in the training data, as shown in Table~\ref{tab:annotate}.
Instead of simple lexical matching with $r \in \boldsymbol{x}$, which is prone to errors in the annotation, 
we use the information extraction system provided by \citet{wiseman2017challenges}.
Although this system is trained on noisy rule-based annotations, 
we conjecture that it is more robust to errors because it is trained to minimize the marginalized loss function for ambiguous relations.
All training details are described in Appendix~\ref{sec:settings}.

\subsection{Models to be compared}
We compare our model\footnote{Our code is available from \url{https://github.com/aistairc/sports-reporter}}
against two baseline models. 
One is the model used by \citet{wiseman2017challenges}, which generates a summary with an attention-based encoder-decoder model.
The other baseline model is the one proposed by \citet{puduppully2019data}, which first predicts the sequence of data records and then generates a summary conditioned on the predicted sequences.
\citet{wiseman2017challenges}'s model refers to all data records every timestep, while \citet{puduppully2019data}'s model refers to a subset of all data records, which is predicted in the first stage.
Unlike these models, our model uses one memory vector $\boldsymbol{h}^{\textsc{Ent}}_{t}$ that tracks the history of the data records, during generation.
We retrained the baselines on our new dataset.
We also present the performance of the \textsc{Gold} and \textsc{Templates} summaries. The \textsc{Gold} summary is exactly identical with the reference summary and each \textsc{Templates} summary is generated in the same manner as \citet{wiseman2017challenges}.

In the latter half of our experiments, 
we examine the effect of adding information about writers.
In addition to our model enhanced with writer information,
we also add writer information to the model by \citet{puduppully2019data}.
Their method consists of two stages corresponding to content planning and surface realization.
Therefore, by incorporating writer information to each of the two stages, we can clearly see which part of the model to which the writer information contributes to. 
For \citet{puduppully2019data} model, we attach the writer information in the following three ways: 
\vspace{-1.em}
\begin{enumerate}
    \setlength{\itemsep}{0pt}
    \setlength{\parskip}{0pt}
     \setlength{\leftskip}{-3mm}
    \item concatenating writer embedding $\boldsymbol{w}$ with the input vector for LSTM in the content planning decoder (stage 1);
    \item concatenating writer embedding $\boldsymbol{w}$ with the input vector for LSTM in the text generator (stage 2);
    \item using both 1 and 2 above.
\end{enumerate}\vspace{-0.5em}
For more details about each decoding stage, readers can refer to \citet{puduppully2019data}.

\subsection{Evaluation metrics}
As evaluation metrics, we use BLEU score~\cite{papineni2002bleu} and the extractive metrics proposed by \citet{wiseman2017challenges}, i.e., relation generation (RG), content selection (CS), and content ordering (CO) as evaluation metrics. 
The extractive metrics measure how well the relations extracted from the generated summary match the correct relations\footnote{The model for extracting relation tuples was trained 
on tuples made from the entity (e.g., team name, city name, player name) and attribute value (e.g., ``Lakers'', ``92'') extracted from the summaries, and the corresponding attributes (e.g., ``\textsc{Team Name}'', ``\textsc{Pts}'') found in the box- or line-score. The precision and the recall of this extraction model are respectively 93.4\% and 75.0\% in the test data.}:

\begin{itemize}
    \setlength{\itemsep}{0pt}
    \setlength{\parskip}{0pt}
    \setlength{\leftskip}{-4mm}
    \item[-] RG: 
	the ratio of the correct relations out of all the extracted relations, where correct relations are relations found in the input data records $\boldsymbol{x}$. The average number of extracted relations is also reported.
	\item[-] CS: precision and recall of the relations extracted from the generated summary against those from the reference summary.
	\item[-] CO: edit distance measured with normalized Damerau-Levenshtein Distance (DLD) between the sequences of relations extracted from the generated and reference summary. 
\end{itemize}
\vspace{-0.5em}

\section{Results and Discussions}
We first focus on the quality of tracking model and entity representation in Sections~\ref{subsec:result_tracking} to
\ref{subsec:result_output}, where we use the model without writer information.
We examine the effect of writer information in Section~\ref{subsec:result_writer}.
\subsection{Saliency tracking-based model}
\label{subsec:result_tracking}
As shown in Table~\ref{tab:result}, our model outperforms all baselines across all evaluation metrics.\footnote{The scores of \citet{puduppully2019data}'s model significantly dropped from what they reported, especially on BLEU metric.
We speculate this is mainly due to the reduced amount of our training data (Section~\ref{section:data}).
That is, their model might be more data-hungry than other models.
}
One of the noticeable results is that our model achieves slightly higher RG precision than the gold summary.
Owing to the extractive evaluation nature, the generated summary of the precision of the relation generation could beat the gold summary performance.
In fact, the template model achieves 100\% precision of the relation generations.

The other is that only our model exceeds the template model regarding F1 score of the content selection and obtains the highest performance of content ordering.
This imply that the tracking model encourages to select salient input records in the correct order.

\begin{figure}[t]
    \centering
    \includegraphics[width=\linewidth]{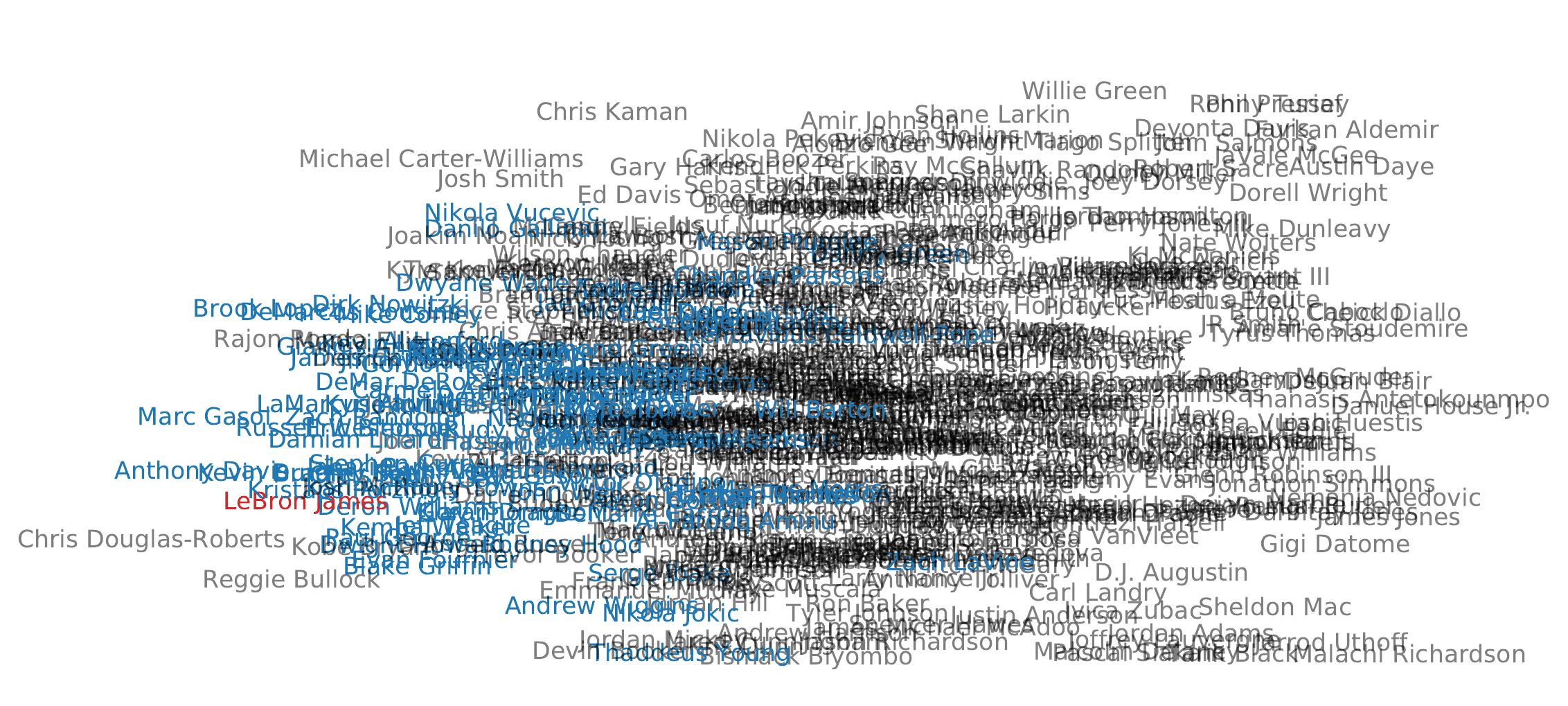}
    \caption{Illustrations of static entity embeddings $\boldsymbol{e}$. Players with colored letters are listed in the ranking top 100 players for the 2016-17 NBA season at \url{https://www.washingtonpost.com/graphics/sports/nba-top-100-players-2016/}. Only {\it LeBron James} is in {\color[HTML]{d62728}red} and the other players in top 100 are in {\color[HTML]{1f77b4}blue}. Top-ranked players have similar representations of $\boldsymbol{e}$.}
    \label{fig:static_ent}
\end{figure}

\begin{figure*}[t]
    \centering
    \includegraphics[width=\linewidth]{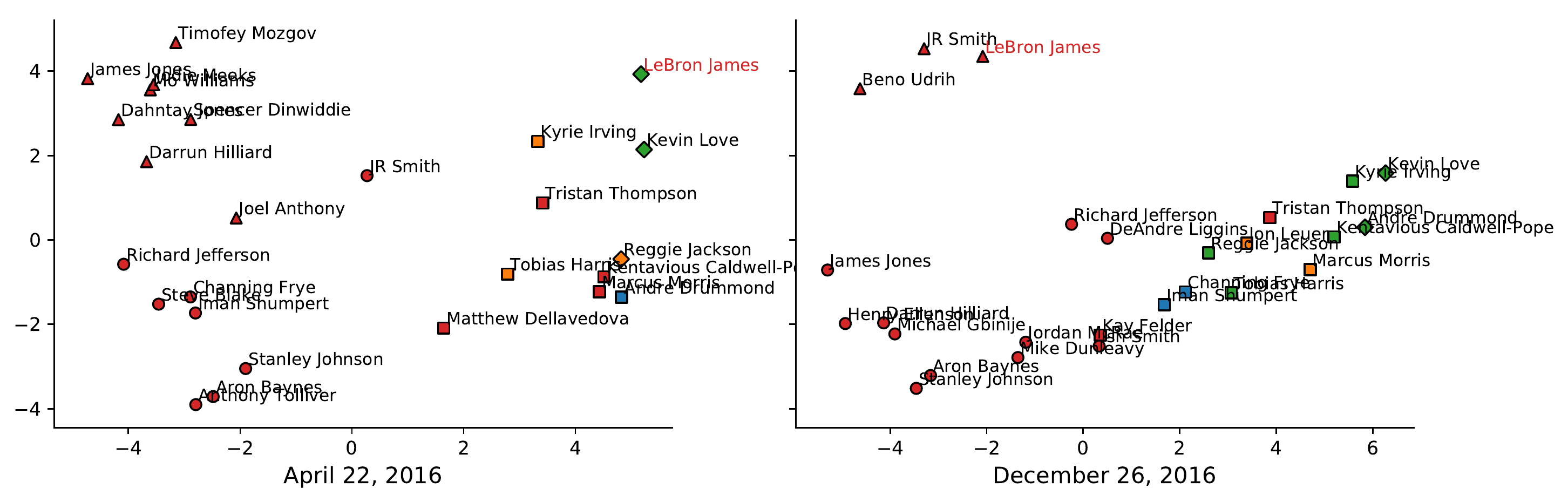}
    \caption{Illustrations of dynamic entity embedding $\bar{\boldsymbol{e}}$. Both left and right figures are for \textit{Cleveland Cavaliers} vs. \textit{Detroit Pistons}, on different dates. LeBron James is in {\color[HTML]{d62728} \textbf{red letters}}.
    {\color[HTML]{ff7f0e}\textbf{Entities with orange symbols}} appeared only in the reference summary.
    {\color[HTML]{1f77b4}\textbf{Entities with blue symbols}} appeared only in the generated summary.
    {\color[HTML]{2ca02c}\textbf{Entities with green symbols}} appeared in both the reference and the generated summary.
    The others are with {\color[HTML]{d62728}\textbf{red symbols}}.
    $\Box$ represents player who scored in the double digits, and $\Diamond$ represents player who recorded double-double. Players with $\triangle$ did not participate in the game. $\circ$ represents other players.
    }
    \label{fig:ebar}
\end{figure*}

\subsection{Qualitative analysis of entity embedding}
\label{subsec:result_entity}
Our model has the entity embedding $\bar{\boldsymbol{e}}$, which depends on the box score for each game in addition to static entity embedding $\boldsymbol{e}$.
Now we analyze the difference of these two types of embeddings.

We present a two-dimensional visualizations of both embeddings produced using PCA~\cite{pearson1901liii}.
As shown in Figure~\ref{fig:static_ent}, which is the visualization of static entity embedding $\boldsymbol{e}$, the top-ranked players are closely located.

We also present the visualizations of dynamic entity embeddings $\bar{\boldsymbol{e}}$ in Figure~\ref{fig:ebar}.
Although we did not carry out feature engineering specific to the NBA (e.g., whether a player scored double digits or not)\footnote{In the NBA, a player who accumulates a \textit{double}-digit score in one of five categories (points, rebounds, assists, steals, and blocked shots) in a game, is regarded as a good player. If a player had a double in two of those five categories, it is referred to as \textit{double-double}.} for representing the dynamic entity embedding $\bar{\boldsymbol{e}}$, the embeddings of the players who performed well for each game have similar representations.
In addition, the change in embeddings of the same player was observed depending on the box-scores for each game.
For instance, \textit{LeBron James} recorded a double-double in a game on April 22, 2016.
For this game, his embedding is located close to the embedding of \textit{Kevin Love}, who also scored a double-double.
However, he did not participate in the game on December 26, 2016. His embedding for this game became closer to those of other players who also did not participate.

\subsection{Duplicate ratios of extracted relations}
\label{subsec:result_duplicate}
As \citet{puduppully2019data} pointed out, 
a generated summary may mention the same relation multiple times. Such duplicated relations are not favorable in terms of the brevity of text.

\begin{figure}[t]
    \centering
    \includegraphics[width=\linewidth]{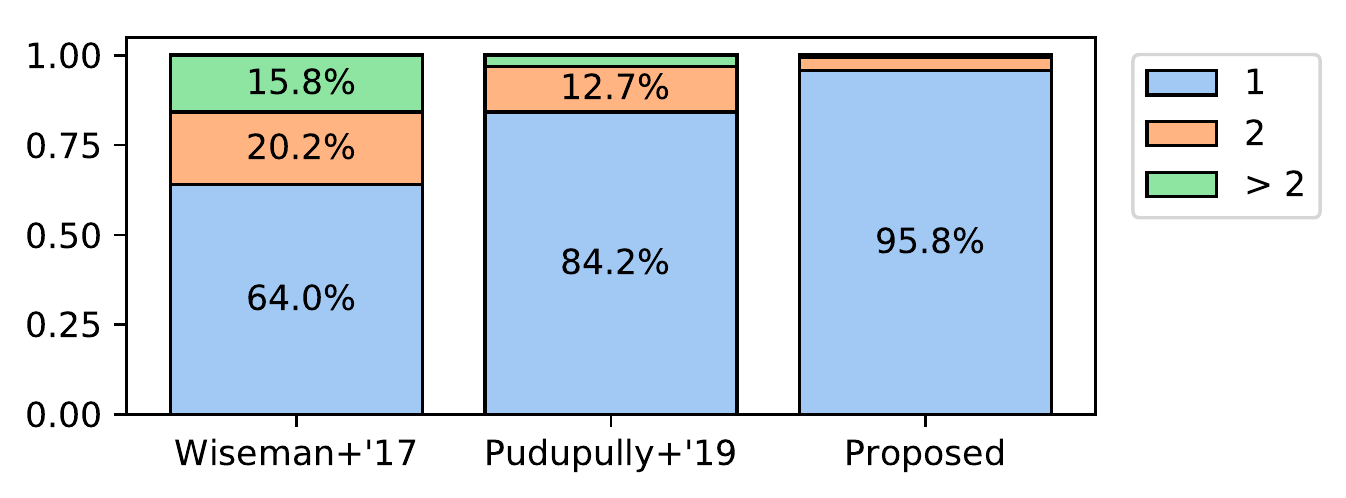}
    \caption{Ratios of generated summaries with duplicate mention of relations. Each label represents number of duplicated relations within each document. While \citet{wiseman2017challenges}'s model exhibited 36.0\% duplication and \citet{puduppully2019data}'s model exhibited 15.8\%, our model exhibited only 4.2\%.}
    \label{fig:ratio}
\end{figure}

Figure~\ref{fig:ratio} shows the ratios of the generated summaries with duplicate mentions of relations in the development data.
While the models by \citet{wiseman2017challenges} and \citet{puduppully2019data} respectively showed 36.0\% and 15.8\% as duplicate ratios, our model exhibited 4.2\%.
This suggests that our model dramatically suppressed generation of redundant relations.
We speculate that the tracking model successfully memorized which input records have been selected in $\boldsymbol{h}_s^{\textsc{Ent}}$.

\subsection{Qualitative analysis of output examples}
\label{subsec:result_output}
Figure~\ref{fig:example} shows the generated examples from validation inputs with \citet{puduppully2019data}'s model and our model.
Whereas both generations seem to be fluent, the summary of \citet{puduppully2019data}'s model includes erroneous relations colored in {\color[HTML]{ff7f0e} orange}.

Specifically, the description about \textsc{Derrick Rose}'s relations, ``15 points, four assists, three rounds and one steal in 33 minutes.'', is also used for other entities (e.g., \textsc{John Henson} and \textsc{Willy Hernagomez}).
This is because \citet{puduppully2019data}'s model has no tracking module unlike our model, which mitigates redundant references and therefore rarely contains erroneous relations.

However, when complicated expressions such as parallel structures are used
our model also generates erroneous relations 
as illustrated by the underlined sentences describing the two players who scored the same points.
For example, ``11-point efforts'' is correct for \textsc{Courtney Lee} but not for \textsc{Derrick Rose}.
As a future study, it is necessary to develop a method that can handle such complicated relations.

\subsection{Use of writer information}
\label{subsec:result_writer}
\begin{table*}[t]{
    \centering
    \addtolength{\tabcolsep}{-2pt}
    \begin{tabular}{l|cc|ccc|c|c}
        \toprule
        \multirow{2}{*}{Method} & \multicolumn{2}{c|}{\textsc{RG}} &\multicolumn{3}{c|}{\textsc{CS}} & \textsc{CO} & \multirow{2}{*}{\textsc{Bleu}}\\
        & \# & P\% & P\% & R\% & F1\% & DLD\%\\
        \midrule\midrule
         \citet{puduppully2019data} &    33.06 & 83.17 & 33.06 & 43.59 & 37.60 & 16.97 & 13.96\\
        + $\boldsymbol{w}$ in stage 1 & 28.43 & \textbf{84.75} & \textbf{45.00} & \textbf{49.73} & \textbf{47.25} & 22.16 & 18.18 \\
        + $\boldsymbol{w}$ in stage 2 & 35.06 & 80.51 & 31.10 & 45.28 & 36.87 & 16.38 & 17.81 \\
        + $\boldsymbol{w}$ in stage 1 \& 2 & 28.00 & 82.27 & 44.37 & 48.71 & 46.44 & \textbf{22.41} & \textbf{18.90} \\
        \midrule
        \textsc{Proposed} & 39.05 & \textbf{94.38} & 35.77 & 52.05 & 42.40 & 19.38 & 16.15\\
        + $\boldsymbol{w}$& 30.25 & 92.00 & \textbf{50.75} & \textbf{59.03} & \textbf{54.58} & \textbf{25.75} & \textbf{20.84} \\
        \bottomrule
    \end{tabular}
    \caption{Effects of writer information. $\boldsymbol{w}$ indicates that \textsc{Writer} embeddings are used. Numbers in \textbf{bold} are the largest among the variants of each method.} 
    \label{tab:author}
    }
\end{table*}

\begin{table*}[t]
    \scriptsize
    \centering
    \begin{subtable}{0.446\linewidth}
    \begin{tikzpicture}
    \node [rectangle, draw, thick, fill=blue!0, text width=0.92\linewidth,  rounded corners, inner sep = 6pt, minimum height=33em]{
    The \textbf{Milwaukee Bucks} defeated the \textbf{New York Knicks}, {\color[HTML]{2ca02c}\textbf{105}}-{\color[HTML]{2ca02c}\textbf{104}}, at Madison Square Garden on Wednesday evening. The \textbf{Bucks} ({\color[HTML]{2ca02c}\textbf{18}}-{\color[HTML]{2ca02c}\textbf{16}}) have been one of the hottest teams in the league, having won five of their last six games, and they have now won six of their last eight games. The \textbf{Knicks} ({\color[HTML]{2ca02c}\textbf{16}}-{\color[HTML]{2ca02c}\textbf{19}}) have now won six of their last six games, as they continue to battle for the eighth and final playoff spot in the Eastern Conference. \textbf{Giannis Antetokounmpo} led the way for Milwaukee, as he tallied {\color[HTML]{2ca02c}\textbf{27}} points, {\color[HTML]{2ca02c}\textbf{13}} rebounds, {\color[HTML]{2ca02c}\textbf{four}} assists, {\color[HTML]{2ca02c}\textbf{three}} blocked shots and {\color[HTML]{1f77b4}\textbf{one}} steal, in {\color[HTML]{1f77b4}\textbf{39}} minutes . \textbf{Jabari Parker} added {\color[HTML]{2ca02c}\textbf{15}} points, {\color[HTML]{2ca02c}\textbf{four}} rebounds, {\color[HTML]{2ca02c}\textbf{three}} assists, {\color[HTML]{1f77b4}\textbf{one}} steal and {\color[HTML]{ff7f0e}\textbf{one}} block, and {\color[HTML]{ff7f0e}\textbf{6}}-of-{\color[HTML]{ff7f0e}\textbf{8}} from long range. \textbf{John Henson} added {\color[HTML]{1f77b4}\textbf{two}} points, {\color[HTML]{1f77b4}\textbf{two}} rebounds, {\color[HTML]{ff7f0e}\textbf{one}} assist, {\color[HTML]{ff7f0e}\textbf{three}} steals and {\color[HTML]{ff7f0e}\textbf{one}} block. \textbf{John Henson} was the only other player to score in double digits for the Knicks, with {\color[HTML]{ff7f0e}\textbf{15}} points, {\color[HTML]{ff7f0e}\textbf{four}} assists, {\color[HTML]{ff7f0e}\textbf{three}} rebounds and {\color[HTML]{ff7f0e}\textbf{one}} steal, in {\color[HTML]{ff7f0e}\textbf{33}} minutes. The Bucks were led by \textbf{Derrick Rose}, who tallied {\color[HTML]{2ca02c}\textbf{15}} points, {\color[HTML]{2ca02c}\textbf{four}} assists, {\color[HTML]{2ca02c}\textbf{three}} rebounds and {\color[HTML]{1f77b4}\textbf{one}} steal in {\color[HTML]{1f77b4}\textbf{33}} minutes. \textbf{Willy Hernangomez} started in place of Porzingis and finished with {\color[HTML]{ff7f0e}\textbf{15}} points, {\color[HTML]{ff7f0e}\textbf{four}} assists, {\color[HTML]{ff7f0e}\textbf{three}} rebounds and {\color[HTML]{ff7f0e}\textbf{one}} steal in {\color[HTML]{ff7f0e}\textbf{33}} minutes. \textbf{Willy Hernangomez} started in place of Jose Calderon ( knee ) and responded with {\color[HTML]{1f77b4}\textbf{one}} rebound and {\color[HTML]{ff7f0e}\textbf{one}} block. The Knicks were led by their starting backcourt of \textbf{Carmelo Anthony} and \textbf{Carmelo Anthony}, but combined for just {\color[HTML]{ff7f0e}\textbf{13}} points on 5-of-16 shooting. The Bucks next head to Philadelphia to take on the Sixers on Friday night, while the Knicks remain home to face the Los Angeles Clippers on Wednesday.
    };
    \end{tikzpicture}
    \caption{\citet{puduppully2019data}}
    \end{subtable}
    \begin{subtable}{0.5\linewidth}
    \begin{tikzpicture}
    \node [rectangle, draw, thick, fill=blue!0, text width=0.92\linewidth,  rounded corners, inner sep = 6pt, minimum height=33em]{
    The \textbf{Milwaukee Bucks} defeated the \textbf{New York Knicks}, {\color[HTML]{2ca02c}\textbf{105}}-{\color[HTML]{2ca02c}\textbf{104}}, at Madison Square Garden on Saturday. The \textbf{Bucks} ({\color[HTML]{2ca02c}\textbf{18}}-{\color[HTML]{2ca02c}\textbf{16}}) checked in to Saturday's contest with a well, outscoring the \textbf{Knicks} ({\color[HTML]{2ca02c}\textbf{16}}-{\color[HTML]{2ca02c}\textbf{19}}) by a margin of 39-19 in the first quarter. However, New York by just a 25-foot lead at the end of the first quarter, the Bucks were able to pull away, as they outscored the Knicks by a 59-46 margin into the second. 45 points in the third quarter to seal the win for New York with the rest of the starters to seal the win. The Knicks were led by \textbf{Giannis Antetokounmpo}, who tallied a game-high {\color[HTML]{2ca02c}\textbf{27}} points, to go along with {\color[HTML]{2ca02c}\textbf{13}} rebounds, {\color[HTML]{2ca02c}\textbf{four}} assists, {\color[HTML]{2ca02c}\textbf{three}} blocks and a steal. The game was a crucial night for the Bucks' starting five, as the duo was the most effective shooters, as they posted Milwaukee to go on a pair of low low-wise (Carmelo Anthony) and Malcolm Brogdon. \textbf{Anthony} added {\color[HTML]{2ca02c}\textbf{11}} rebounds, {\color[HTML]{2ca02c}\textbf{seven}} assists and {\color[HTML]{2ca02c}\textbf{two}} steals to his team-high scoring total. \textbf{Jabari Parker} was right behind him with {\color[HTML]{2ca02c}\textbf{15}} points, {\color[HTML]{2ca02c}\textbf{four}} rebounds, {\color[HTML]{2ca02c}\textbf{three}} assists and a block. \textbf{Greg Monroe} was next with a bench-leading {\color[HTML]{2ca02c}\textbf{18}} points, along with {\color[HTML]{2ca02c}\textbf{nine}} rebounds, {\color[HTML]{2ca02c}\textbf{four}} assists and {\color[HTML]{2ca02c}\textbf{three}} steals. \textbf{Brogdon} posted {\color[HTML]{2ca02c}\textbf{12}} points, {\color[HTML]{2ca02c}\textbf{eight}} assists, {\color[HTML]{2ca02c}\textbf{six}} rebounds and a steal. \underline{\textbf{Derrick Rose} and \textbf{Courtney Lee} were next with a pair of \{{\color[HTML]{ff7f0e}\textbf{11}}~/~{\color[HTML]{2ca02c}\textbf{11}}\}} \underline{-point efforts.} \textbf{Rose} also supplied {\color[HTML]{2ca02c}\textbf{four}} assists and {\color[HTML]{2ca02c}\textbf{three}} rebounds, while \textbf{Lee} complemented his scoring with {\color[HTML]{2ca02c}\textbf{three}} assists, a rebound and a steal. \underline{\textbf{John Henson} and \textbf{Mirza Teletovic} were next with a pair of \{{\color[HTML]{1f77b4}\textbf{two}}~/~{\color[HTML]{ff7f0e}\textbf{two}}\}} \underline{-point efforts.} \textbf{Teletovic} also registered {\color[HTML]{2ca02c}\textbf{13}} points, and he added a rebound and an assist. \textbf{Jason Terry} supplied {\color[HTML]{1f77b4}\textbf{eight}} points, {\color[HTML]{1f77b4}\textbf{three}} rebounds and a pair of steals. The Cavs remain in last place in the Eastern Conference's Atlantic Division. They now head home to face the Toronto Raptors on Saturday night.
    };
    \end{tikzpicture}
    \caption{Our model}
    \end{subtable}
    \caption{Example summaries generated with \citet{puduppully2019data}'s model (left) and our model (right). Names in \textbf{bold face} are salient entities. {\color[HTML]{1f77b4}\textbf{Blue numbers}} are correct relations derived from input data records but are not observed in reference summary.
    {\color[HTML]{ff7f0e}\textbf{Orange numbers}} are incorrect relations.
    {\color[HTML]{2ca02c}\textbf{Green numbers}} are correct relations mentioned in reference summary.
    }
    \label{fig:example}
\end{table*}
We first look at the results of an extension of \citet{puduppully2019data}'s model with writer information $\boldsymbol{w}$
in Table~\ref{tab:author}.
By adding $\boldsymbol{w}$ to content planning (stage 1), the method obtained improvements in CS (37.60 to 47.25), CO (16.97 to 22.16), and BLEU score (13.96 to 18.18).
By adding $\boldsymbol{w}$ to the component for surface realization (stage 2), the method obtained an improvement in BLEU score (13.96 to 17.81), while the effects on the other metrics were not very significant.
By adding $\boldsymbol{w}$ to both stages, the method scored the highest BLEU, while the other metrics were not very different from those obtained by adding $\boldsymbol{w}$ to stage 1.
This result suggests that writer information contributes to both content planning and surface realization when it is properly used, and improvements of content planning lead to much better performance in surface realization.

Our model showed improvements in most metrics and showed the best performance by incorporating writer information $\boldsymbol{w}$.
As discussed in Section~\ref{subsec:writer},
$\boldsymbol{w}$ is supposed to affect both content planning and surface realization. Our experimental result is consistent with the discussion. 

\section{Conclusion}
In this research, we proposed a new data-to-text model that produces a summary text while tracking the salient information that imitates a human-writing process.
As a result, our model outperformed the existing models in all evaluation measures.
We also explored the effects of incorporating writer information to data-to-text models.
With writer information, our model successfully generated highest quality summaries that scored 20.84 points of \textsc{BLEU} score.

\section*{Acknowledgments}
We would like to thank the anonymous reviewers for
their helpful suggestions. 
This paper is based on results obtained from a
project commissioned by the New Energy and
Industrial Technology Development Organization
(NEDO), JST PRESTO (Grant Number JPMJPR1655), and AIST-Tokyo Tech Real World Big-Data Computation Open Innovation Laboratory (RWBC-OIL).

\bibliography{acl2019}

\begin{thebibliography}{36}
\expandafter\ifx\csname natexlab\endcsname\relax\def\natexlab#1{#1}\fi

\bibitem[{Aoki et~al.(2018)Aoki, Miyazawa, Ishigaki, Goshima, Aoki, Kobayashi,
  Takamura, and Miyao}]{aoki2018generating}
Tatsuya Aoki, Akira Miyazawa, Tatsuya Ishigaki, Keiichi Goshima, Kasumi Aoki,
  Ichiro Kobayashi, Hiroya Takamura, and Yusuke Miyao. 2018.
\newblock \href {http://aclweb.org/anthology/W18-6515} {{Generating Market
  Comments Referring to External Resources}}.
\newblock In \emph{Proceedings of the 11th International Conference on Natural
  Language Generation}, pages 135--139.

\bibitem[{Bahdanau et~al.(2015)Bahdanau, Cho, and Bengio}]{bahdanau2015neural}
Dzmitry Bahdanau, Kyunghyun Cho, and Yoshua Bengio. 2015.
\newblock \href {https://arxiv.org/pdf/1409.0473.pdf} {{Neural machine
  translation by jointly learning to align and translate}}.
\newblock In \emph{Proceedings of the Third International Conference on
  Learning Representations}.

\bibitem[{Barzilay and Lapata(2005)}]{barzilay2005collective}
Regina Barzilay and Mirella Lapata. 2005.
\newblock \href {http://aclweb.org/anthology/H05-1042} {{Collective content
  selection for concept-to-text generation}}.
\newblock In \emph{Proceedings of the conference on Human Language Technology
  and Empirical Methods in Natural Language Processing}, pages 331--338.

\bibitem[{Bosselut et~al.(2018)Bosselut, Levy, Holtzman, Ennis, Fox, and
  Choi}]{bosselut2018simulating}
Antoine Bosselut, Omer Levy, Ari Holtzman, Corin Ennis, Dieter Fox, and Yejin
  Choi. 2018.
\newblock \href {https://openreview.net/pdf?id=rJYFzMZC-} {{Simulating Action
  Dynamics with Neural Process Networks}}.
\newblock In \emph{Proceedings of the Sixth International Conference on
  Learning Representations}.

\bibitem[{Chen and Mooney(2008)}]{chen2008learning}
David~L Chen and Raymond~J Mooney. 2008.
\newblock \href {https://icml.cc/Conferences/2008/papers/304.pdf} {{Learning to
  sportscast: a test of grounded language acquisition}}.
\newblock In \emph{Proceedings of the 25th international conference on Machine
  learning}, pages 128--135.

\bibitem[{Cho et~al.(2014)Cho, van Merrienboer, Gulcehre, Bahdanau, Bougares,
  Schwenk, and Bengio}]{cho2014learning}
Kyunghyun Cho, Bart van Merrienboer, Caglar Gulcehre, Dzmitry Bahdanau, Fethi
  Bougares, Holger Schwenk, and Yoshua Bengio. 2014.
\newblock \href {https://www.aclweb.org/anthology/D14-1179} {{Learning Phrase
  Representations using RNN Encoder--Decoder for Statistical Machine
  Translation}}.
\newblock In \emph{Proceedings of the 2014 Conference on Empirical Methods in
  Natural Language Processing}, pages 1724--1734.

\bibitem[{Chung et~al.(2014)Chung, Gulcehre, Cho, and
  Bengio}]{chung2014empirical}
Junyoung Chung, Caglar Gulcehre, KyungHyun Cho, and Yoshua Bengio. 2014.
\newblock {Empirical evaluation of gated recurrent neural networks on sequence
  modeling}.
\newblock \emph{arXiv preprint arXiv:1412.3555}.

\bibitem[{Clark et~al.(2018)Clark, Ji, and Smith}]{clark2018neural}
Elizabeth Clark, Yangfeng Ji, and Noah~A Smith. 2018.
\newblock \href {http://aclweb.org/anthology/N18-1204} {{Neural Text Generation
  in Stories Using Entity Representations as Context}}.
\newblock In \emph{Proceedings of the 16th Conference of the North American
  Chapter of the Association for Computational Linguistics: Human Language
  Technologies}, pages 2250--2260.

\bibitem[{Glorot and Bengio(2010)}]{glorot2010understanding}
Xavier Glorot and Yoshua Bengio. 2010.
\newblock \href {http://proceedings.mlr.press/v9/glorot10a/glorot10a.pdf}
  {{Understanding the difficulty of training deep feedforward neural
  networks}}.
\newblock In \emph{Proceedings of the thirteenth international conference on
  artificial intelligence and statistics}, pages 249--256.

\bibitem[{Graves et~al.(2016)Graves, Wayne, Reynolds, Harley, Danihelka,
  Grabska-Barwi{\'n}ska, Colmenarejo, Grefenstette, Ramalho, Agapiou
  et~al.}]{graves2016hybrid}
Alex Graves, Greg Wayne, Malcolm Reynolds, Tim Harley, Ivo Danihelka, Agnieszka
  Grabska-Barwi{\'n}ska, Sergio~G{\'o}mez Colmenarejo, Edward Grefenstette,
  Tiago Ramalho, John Agapiou, et~al. 2016.
\newblock \href {https://www.nature.com/articles/nature20101.pdf} {{Hybrid
  computing using a neural network with dynamic external memory}}.
\newblock \emph{Nature}, 538(7626):471.

\bibitem[{Gu et~al.(2016)Gu, Lu, Li, and Li}]{gu2016incorporating}
Jiatao Gu, Zhengdong Lu, Hang Li, and Victor~OK Li. 2016.
\newblock \href {http://www.aclweb.org/anthology/P16-1154} {{Incorporating
  Copying Mechanism in Sequence-to-Sequence Learning}}.
\newblock In \emph{Proceedings of the 54th Annual Meeting of the Association
  for Computational Linguistics}, volume~1, pages 1631--1640.

\bibitem[{Gulcehre et~al.(2016)Gulcehre, Ahn, Nallapati, Zhou, and
  Bengio}]{gulcehre2016pointing}
Caglar Gulcehre, Sungjin Ahn, Ramesh Nallapati, Bowen Zhou, and Yoshua Bengio.
  2016.
\newblock \href {http://aclweb.org/anthology/P16-1014} {{Pointing the Unknown
  Words}}.
\newblock In \emph{Proceedings of the 54th Annual Meeting of the Association
  for Computational Linguistics}, pages 140--149.

\bibitem[{Hoang et~al.(2018)Hoang, Wiseman, and Rush}]{hoang2018entity}
Luong Hoang, Sam Wiseman, and Alexander Rush. 2018.
\newblock \href {http://www.aclweb.org/anthology/D18-1130} {{Entity Tracking
  Improves Cloze-style Reading Comprehension}}.
\newblock In \emph{Proceedings of the 2018 Conference on Empirical Methods in
  Natural Language Processing}, pages 1049--1055.

\bibitem[{Ji et~al.(2017)Ji, Tan, Martschat, Choi, and Smith}]{ji2017dynamic}
Yangfeng Ji, Chenhao Tan, Sebastian Martschat, Yejin Choi, and Noah~A Smith.
  2017.
\newblock \href {http://aclweb.org/anthology/D17-1195} {{Dynamic Entity
  Representations in Neural Language Models}}.
\newblock In \emph{Proceedings of the 2017 Conference on Empirical Methods in
  Natural Language Processing}, pages 1830--1839.

\bibitem[{Kiddon et~al.(2016)Kiddon, Zettlemoyer, and
  Choi}]{kiddon2016globally}
Chlo{\'e} Kiddon, Luke Zettlemoyer, and Yejin Choi. 2016.
\newblock \href {http://aclweb.org/anthology/D16-1032} {{Globally coherent text
  generation with neural checklist models}}.
\newblock In \emph{Proceedings of the 2016 Conference on Empirical Methods in
  Natural Language Processing}, pages 329--339.

\bibitem[{Kobayashi et~al.(2016)Kobayashi, Tian, Okazaki, and
  Inui}]{kobayashi2016dynamic}
Sosuke Kobayashi, Ran Tian, Naoaki Okazaki, and Kentaro Inui. 2016.
\newblock \href {http://www.aclweb.org/anthology/N16-1099} {{Dynamic entity
  representation with max-pooling improves machine reading}}.
\newblock In \emph{Proceedings of the 15th Conference of the North American
  Chapter of the Association for Computational Linguistics: Human Language
  Technologies}, pages 850--855.

\bibitem[{Lebret et~al.(2016)Lebret, Grangier, and Auli}]{lebret2016neural}
R{\'e}mi Lebret, David Grangier, and Michael Auli. 2016.
\newblock \href {http://www.aclweb.org/anthology/D16-1128} {{Neural Text
  Generation from Structured Data with Application to the Biography Domain}}.
\newblock In \emph{Proceedings of the 2016 Conference on Empirical Methods in
  Natural Language Processing}, pages 1203--1213.

\bibitem[{Liang et~al.(2009)Liang, Jordan, and Klein}]{liang2009learning}
Percy Liang, Michael~I Jordan, and Dan Klein. 2009.
\newblock \href {http://aclweb.org/anthology/P09-1011} {{Learning semantic
  correspondences with less supervision}}.
\newblock In \emph{Proceedings of the Joint Conference of the 47th Annual
  Meeting of the ACL and the 4th International Joint Conference on Natural
  Language Processing of the AFNLP}, pages 91--99.

\bibitem[{Liu et~al.(2018)Liu, Wang, Sha, Chang, and Sui}]{liu2018table}
Tianyu Liu, Kexiang Wang, Lei Sha, Baobao Chang, and Zhifang Sui. 2018.
\newblock \href
  {https://www.aaai.org/ocs/index.php/AAAI/AAAI18/paper/viewFile/16599/16019}
  {{Table-to-text Generation by Structure-aware Seq2seq Learning}}.
\newblock In \emph{Proceedings of the Thirty-Second AAAI Conference on
  Artificial Intelligence}.

\bibitem[{Luong et~al.(2015)Luong, Pham, and Manning}]{luong2015effective}
Thang Luong, Hieu Pham, and Christopher~D Manning. 2015.
\newblock \href {http://www.aclweb.org/anthology/D15-1166} {{Effective
  Approaches to Attention-based Neural Machine Translation}}.
\newblock In \emph{Proceedings of the 2015 Conference on Empirical Methods in
  Natural Language Processing}, pages 1412--1421.

\bibitem[{Mei et~al.(2016)Mei, Bansal, and Walter}]{mei2016talk}
Hongyuan Mei, Mohit Bansal, and Matthew~R Walter. 2016.
\newblock \href {http://www.aclweb.org/anthology/N16-1086} {{What to talk about
  and how? Selective Generation using LSTMs with Coarse-to-Fine Alignment}}.
\newblock In \emph{Proceedings of the 15th Conference of the North American
  Chapter of the Association for Computational Linguistics: Human Language
  Technologies}, pages 720--730.

\bibitem[{Murakami et~al.(2017)Murakami, Watanabe, Miyazawa, Goshima, Yanase,
  Takamura, and Miyao}]{murakami2017learning}
Soichiro Murakami, Akihiko Watanabe, Akira Miyazawa, Keiichi Goshima, Toshihiko
  Yanase, Hiroya Takamura, and Yusuke Miyao. 2017.
\newblock \href {http://aclweb.org/anthology/P17-1126} {{Learning to generate
  market comments from stock prices}}.
\newblock In \emph{Proceedings of the 55th Annual Meeting of the Association
  for Computational Linguistics}, pages 1374--1384.

\bibitem[{Neubig et~al.(2017)Neubig, Dyer, Goldberg, Matthews, Ammar,
  Anastasopoulos, Ballesteros, Chiang, Clothiaux, Cohn
  et~al.}]{neubig2017dynet}
Graham Neubig, Chris Dyer, Yoav Goldberg, Austin Matthews, Waleed Ammar,
  Antonios Anastasopoulos, Miguel Ballesteros, David Chiang, Daniel Clothiaux,
  Trevor Cohn, et~al. 2017.
\newblock Dynet: The dynamic neural network toolkit.
\newblock \emph{arXiv preprint arXiv:1701.03980}.

\bibitem[{Papineni et~al.(2002)Papineni, Roukos, Ward, and
  Zhu}]{papineni2002bleu}
Kishore Papineni, Salim Roukos, Todd Ward, and Wei-Jing Zhu. 2002.
\newblock \href {http://www.aclweb.org/anthology/P02-1040} {{BLEU: a method for
  automatic evaluation of machine translation}}.
\newblock In \emph{Proceedings of the 40th annual meeting on association for
  computational linguistics}, pages 311--318.

\bibitem[{Pearson(1901)}]{pearson1901liii}
Karl Pearson. 1901.
\newblock {On lines and planes of closest fit to systems of points in space}.
\newblock \emph{The London, Edinburgh, and Dublin Philosophical Magazine and
  Journal of Science}, 2(11):559--572.

\bibitem[{Puduppully et~al.(2019)Puduppully, Dong, and
  Lapata}]{puduppully2019data}
Ratish Puduppully, Li~Dong, and Mirella Lapata. 2019.
\newblock \href
  {https://www.aaai.org/Papers/AAAI/2019/AAAI-PuduppullyR.754.pdf}
  {{{Data-to-Text Generation with Content Selection and Planning}}}.
\newblock In \emph{Proceedings of the Thirty-Third AAAI Conference on
  Artificial Intelligence}.

\bibitem[{Reddi et~al.(2018)Reddi, Kale, and Kumar}]{reddi2018convergence}
Sashank~J Reddi, Satyen Kale, and Sanjiv Kumar. 2018.
\newblock \href {https://openreview.net/pdf?id=ryQu7f-RZ} {{On the convergence
  of adam and beyond}}.
\newblock In \emph{Proceedings of the Sixth International Conference on
  Learning Representations}.

\bibitem[{Sha et~al.(2018)Sha, Mou, Liu, Poupart, Li, Chang, and
  Sui}]{sha2018order}
Lei Sha, Lili Mou, Tianyu Liu, Pascal Poupart, Sujian Li, Baobao Chang, and
  Zhifang Sui. 2018.
\newblock \href
  {https://www.aaai.org/ocs/index.php/AAAI/AAAI18/paper/viewFile/16203/16095}
  {{Order-planning neural text generation from structured data}}.
\newblock In \emph{Proceedings of the Thirty-Second AAAI Conference on
  Artificial Intelligence}.

\bibitem[{Sukhbaatar et~al.(2015)Sukhbaatar, Weston, Fergus
  et~al.}]{sukhbaatar2015end}
Sainbayar Sukhbaatar, Jason Weston, Rob Fergus, et~al. 2015.
\newblock \href
  {https://papers.nips.cc/paper/5846-end-to-end-memory-networks.pdf}
  {{End-to-end memory networks}}.
\newblock In \emph{Advances in neural information processing systems}, pages
  2440--2448.

\bibitem[{Sutskever et~al.(2014)Sutskever, Vinyals, and
  Le}]{sutskever2014sequence}
Ilya Sutskever, Oriol Vinyals, and Quoc~V Le. 2014.
\newblock \href
  {http://papers.nips.cc/paper/5346-sequence-to-sequence-learning-with-neural-networks.pdf}
  {{Sequence to sequence learning with neural networks}}.
\newblock In \emph{Advances in neural information processing systems}, pages
  3104--3112.

\bibitem[{Tanaka-Ishii et~al.(1998)Tanaka-Ishii, Hasida, and
  Noda}]{tanaka1998reactive}
Kumiko Tanaka-Ishii, K{\^o}iti Hasida, and Itsuki Noda. 1998.
\newblock \href {http://aclweb.org/anthology/P98-2209} {{Reactive content
  selection in the generation of real-time soccer commentary}}.
\newblock In \emph{Proceedings of the 36th Annual Meeting of the Association
  for Computational Linguistics and 17th International Conference on
  Computational Linguistics}, pages 1282--1288.

\bibitem[{Taniguchi et~al.(2019)Taniguchi, Feng, Takamura, and
  Okumura}]{taniguchi2019generating}
Yasufumi Taniguchi, Yukun Feng, Hiroya Takamura, and Manabu Okumura. 2019.
\newblock \href
  {https://www.aaai.org/Papers/AAAI/2019/AAAI-TaniguchiY.3507.pdf}
  {{{Generating Live Soccer-Match Commentary from Play Data}}}.
\newblock In \emph{Proceedings of the Thirty-Third AAAI Conference on
  Artificial Intelligence}.

\bibitem[{Tu et~al.(2017)Tu, Liu, Lu, Liu, and Li}]{tu2017context}
Zhaopeng Tu, Yang Liu, Zhengdong Lu, Xiaohua Liu, and Hang Li. 2017.
\newblock \href {http://www.aclweb.org/anthology/Q17-1007} {{Context gates for
  neural machine translation}}.
\newblock \emph{Transactions of the Association for Computational Linguistics},
  5:87--99.

\bibitem[{Weston et~al.(2015)Weston, Chopra, and Bordes}]{weston2015memory}
Jason Weston, Sumit Chopra, and Antoine Bordes. 2015.
\newblock \href {https://arxiv.org/pdf/1410.3916.pdf} {{Memory Networks}}.
\newblock In \emph{Proceedings of the Third International Conference on
  Learning Representations}.

\bibitem[{Wiseman et~al.(2017)Wiseman, Shieber, and
  Rush}]{wiseman2017challenges}
Sam Wiseman, Stuart Shieber, and Alexander Rush. 2017.
\newblock \href {http://aclweb.org/anthology/D17-1239} {{{Challenges in
  Data-to-Document Generation}}}.
\newblock In \emph{Proceedings of the 2017 Conference on Empirical Methods in
  Natural Language Processing}, pages 2253--2263.

\bibitem[{Yang et~al.(2017)Yang, Blunsom, Dyer, and Ling}]{yang2017reference}
Zichao Yang, Phil Blunsom, Chris Dyer, and Wang Ling. 2017.
\newblock \href {https://www.aclweb.org/anthology/D17-1197} {{Reference-Aware
  Language Models}}.
\newblock In \emph{Proceedings of the 2017 Conference on Empirical Methods in
  Natural Language Processing}, pages 1850--1859.

\end{thebibliography}
\bibliographystyle{acl_natbib}

\appendix
\section{Algorithm}
\label{sec:algorithm}
The generation process of our model is shown in Algorithm~\ref{alg}.
For a concise description, we omit the condition for each probability notation. $<$\textsc{SoD}$>$ and $<$\textsc{EoD}$>$ represent ``start of the document'' and ``end of the document'', respectively.
\begin{algorithm*}[b]
    \caption{Generation process}
    \label{alg}
    \KwIn{Data records $\boldsymbol{s}$,\\ Annotations $Z_{1:T}, E_{1:T}, A_{1:T}, N_{1:T}$}
    Initialize $\{\boldsymbol{r}_{e, a, v}\}_{r \in \boldsymbol{x}}$,
    $\{\bar{\boldsymbol{e}}\}_{e\in\mathcal{E}}$, $\boldsymbol{h}_0^{\textsc{LM}}$, $\boldsymbol{h}_0^{\textsc{Ent}}$\\
    $t \leftarrow 0$\\
    $e_t, y_t \leftarrow \textsc{None}, <\textsc{SoD}>$ \\
    \While{$y_t\neq <\textsc{EoD}>$}{
        $t \leftarrow t + 1$\\
        \eIf{$p(Z_t = 1) \geq 0.5$ }{
            \tcc{Select the entity}
            $e_t \leftarrow \arg\max p(E_t=e_t')$ \\
            \uIf{$e_t \not\in \mathcal{E}_{t-1}$}{
                \tcc{If $e_t$ is a new entity}
              $\boldsymbol{h}_{t}^{\textsc{Ent}'} \leftarrow \textsc{Gru}^{\textsc{E}}(\bar{\boldsymbol{e}}_t, \boldsymbol{h}_{t-1}^{\textsc{Ent}})$\\
              $\mathcal{E}_t \leftarrow \mathcal{E}_{t-1} \cup \{e_t\}$\\
               
            }
            \uElseIf{$e_t \neq e_{t-1}$}{
                \tcc{If $e_t$ has been observed before, but is different from the previous one.}
              $\boldsymbol{h}_{t}^{\textsc{Ent'}}\leftarrow \textsc{Gru}^{\textsc{E}}(\boldsymbol{W}^S \boldsymbol{h}_{s}^{\textsc{Ent}},   \boldsymbol{h}_{t-1}^{\textsc{Ent}})$, \\
                where $s = {\max\{s: s\leq t-1, e = e_s\}}$ 
            }
            \uElse{
                $\boldsymbol{h}_{t}^{\textsc{Ent'}}\leftarrow\boldsymbol{h}_{t-1}^{\textsc{Ent}}$
            }
            \tcc{Select an attribute for the entity, $e_t$.}
            $a_t \leftarrow \arg\max p(A_t=a_t')$ \\
            $\boldsymbol{h}_{t}^{\textsc{Ent}} \leftarrow \textsc{Gru}^{\textsc{A}}(\boldsymbol{r}_{e_t, a_t, \boldsymbol{x}[e_t, a_t]}, \boldsymbol{h}_{t}^{\textsc{Ent}'})$\\
            \uIf{$a_t$ \mbox{is a number attribute}}{
                \eIf{$p(N_t=1) \geq 0.5$}{
                    $y_t \leftarrow \text{numeral of } \boldsymbol{x}[e_t,a_t]$
                }{
                    $y_t \leftarrow \boldsymbol{x}[e_t,a_t]$
                }
            }\uElse{
                $y_t \leftarrow \boldsymbol{x}[e_t,a_t]$
            }
            $\boldsymbol{h}_t'\leftarrow \tanh\left(\boldsymbol{W}^{H}(\boldsymbol{h}_{t-1}^{\textsc{LM}} \oplus \boldsymbol{h}_{t}^{\textsc{Ent}})\right)$\\
            $\boldsymbol{h}_t^{\textsc{LM}} \leftarrow \textsc{Lstm}(\boldsymbol{y}_t \oplus \boldsymbol{h}_t', \boldsymbol{h}_{t-1}^{\textsc{LM}})$
        }{
            $e_t, a_t, \boldsymbol{h}_{t}^{\textsc{Ent}} \leftarrow e_{t-1}, a_{t-1}, \boldsymbol{h}_{t-1}^{\textsc{Ent}}$\\
            $\boldsymbol{h}_t'\leftarrow \tanh\left(\boldsymbol{W}^{H}(\boldsymbol{h}_{t-1}^{\textsc{LM}} \oplus \boldsymbol{h}_{t}^{\textsc{Ent}})\right)$\\
            $y_t \leftarrow \arg\max p(Y_t)$ \\
            $\boldsymbol{h}_t^{\textsc{LM}} \leftarrow \textsc{Lstm}(\boldsymbol{y}_t \oplus \boldsymbol{h}_t', \boldsymbol{h}_{t-1}^{\textsc{LM}})$
        }
        \uIf{
            $y_t$ is ``.''
        }{
            $\boldsymbol{h}_t^{\textsc{Ent}} \leftarrow \textsc{Gru}^\textsc{A}(\boldsymbol{v}_{\textsc{Refresh}}, \boldsymbol{h}_t^{\textsc{Ent}})$
        }
    }
    \Return $y_{1:t-1}$;
    
\end{algorithm*}

\section{Experimental settings}
\label{sec:settings}
We set the dimensions of the embeddings to 128, and those of the hidden state of RNN to 512 and all of parameters are initialized with the Xavier initialization~\cite{glorot2010understanding}.
We set the maximum number of epochs to 30, and choose the model with the highest \textsc{Bleu} score on the development data.
The initial learning rate is 2e-3 and AMSGrad is also used for automatically adjusting the learning rate~\cite{reddi2018convergence}.
Our implementation uses DyNet~\cite{neubig2017dynet}.
\end{document}